\documentclass{article}

\usepackage{microtype}
\usepackage{graphicx}
\usepackage{subcaption}
\usepackage{booktabs}
\usepackage{multirow}
\usepackage{xcolor}
\usepackage{hyperref}
\usepackage{xurl}
\usepackage{tikz}
\usetikzlibrary{arrows.meta,positioning,fit,backgrounds}

\usepackage[preprint]{icml2026}

\usepackage{amsmath}
\usepackage{amssymb}
\usepackage{mathtools}
\usepackage[capitalize,noabbrev]{cleveref}

\newcommand{\sd}[1]{\,{\scriptsize$\pm$#1}}

\icmltitlerunning{RainCheck: Harnessing Historical Weather Data for Rainfall Forecasting in Bangladesh}

\begin{document}

\twocolumn[
  \icmltitle{RainCheck: Harnessing Historical Weather Data with Machine Learning\\
    for Rainfall Forecasting in Bangladesh}

  \begin{icmlauthorlist}
    \icmlauthor{Shahran Rahman Alve}{nsu}
    \icmlauthor{Muhammad Zawad Mahmud}{nsu}
    \icmlauthor{Samiha Islam}{nsu}
    \icmlauthor{Al Jubayer Pial}{nsu}
  \end{icmlauthorlist}

  \icmlaffiliation{nsu}{Department of Electrical and Computer Engineering, North South University, Bashundhara, Dhaka-1229, Bangladesh}

  \icmlcorrespondingauthor{Shahran Rahman Alve}{shahran.alve@northsouth.edu}
  \icmlcorrespondingauthor{Muhammad Zawad Mahmud}{zawad.mahmud1@northsouth.edu}
  \icmlcorrespondingauthor{Samiha Islam}{samiha.islam2@northsouth.edu}
\icmlcorrespondingauthor{Al Jubayer Pial}{jubayer.pial@northsouth.edu}

  \icmlkeywords{rainfall forecasting, evaluation, data leakage, time series, climatology.}

  \vskip 0.3in
]

\printAffiliationsAndNotice{}

\begin{abstract}
Machine-learning rainfall forecasts are often evaluated by shuffling station records into random training and test sets. For rainfall, this lets a model see what happened at neighbouring stations on the very day it is asked to forecast, and it hides the fact that a seasonal lookup table already explains most of the variance. We quantify both effects on 618,650 station-days from 35 Bangladesh Meteorological Department stations (1948--2014). We compare random, temporal (train $\le$1999, test 2005--2014) and spatio-temporal (unseen stations and unseen years) evaluation, calendar-only versus antecedent-rainfall features, and six learners against climatology, persistence and SARIMA. Three findings stand out. First, a station-month climatology alone classifies high-rainfall months with 88.4\% accuracy (AUC 0.950), and no learner improved on it by more than 0.8 points under temporal evaluation. Second, a calendar-only random forest reached a next-day wet-day AUC of 0.889 under a random split, higher than any model with real predictors under temporal evaluation, yet fell to 0.835, below climatology (0.859), on future years. Third, models with lagged rainfall add genuine but modest skill for next-day forecasts: a Brier skill score of 0.074 (95\% CI 0.064--0.088) over climatology, rising to 0.091 when station and year inputs are removed and to 0.131 when other stations' same-day observations are added. A GRU sequence model matches but does not exceed gradient boosting, and skill disappears beyond a two-day lead. For monthly totals, no learner beat climatology significantly. We release a leakage-aware protocol and code so that future station-based studies report skill against climatology on held-out years.
\end{abstract}

\section{Introduction}
\label{sec:intro}

Bangladesh receives most of its annual rainfall during the June--September monsoon, and both excess and deficit rainfall have large consequences for floods, agriculture and public health \citep{shahid2010rainfall,karim2020climate}. Station records from the Bangladesh Meteorological Department (BMD) are freely available for several decades \citep{redikod_kaggle}, and studies have trained standard learners such as linear models, $k$-nearest neighbours, decision trees, random forests and support vector machines on such records. Comparable studies in other countries report high accuracies for daily or monthly rainfall \citep{cramer2017extensive,ridwan2021rainfall,pham2020daily,liyew2021daily,barrera2022rainfall,rahman2022fusion}.

These numbers are hard to interpret for two reasons. The first is the \emph{evaluation split}. When station-days are shuffled into training and test sets, the training set contains the same calendar day at other stations. Because rainfall is spatially correlated (the median correlation of same-day rainfall between two BMD stations is 0.31, and 0.51 for rain/no-rain occurrence in our data), a model that only sees the station identity and the date can recover part of the answer from neighbouring stations. This is a form of data leakage \citep{kaufman2012leakage,kapoor2023leakage}, and random cross-validation is known to overstate skill for temporally and spatially structured data \citep{bergmeir2012use,roberts2017cross,ploton2020spatial,cerqueira2020evaluating}. The second reason is the \emph{reference forecast}. Rainfall in Bangladesh has a strong seasonal cycle, so a large share of the variance in monthly totals, and much of the separation between ``high'' and ``low'' rainfall cases, can be obtained by simply forecasting the long-term average for that station and time of year. Forecast verification therefore reports skill \emph{relative to} climatology rather than raw accuracy \citep{murphy1988skill,murphy1993good,wilks2019statistical}.

A common setup trains learners on calendar inputs alone (station, year, month, day), evaluates them on a random split, and reports accuracy or $R^2$ without a climatological reference. In this paper we ask how much of the performance obtained in this way survives a stricter evaluation, and what skill station data can actually provide.

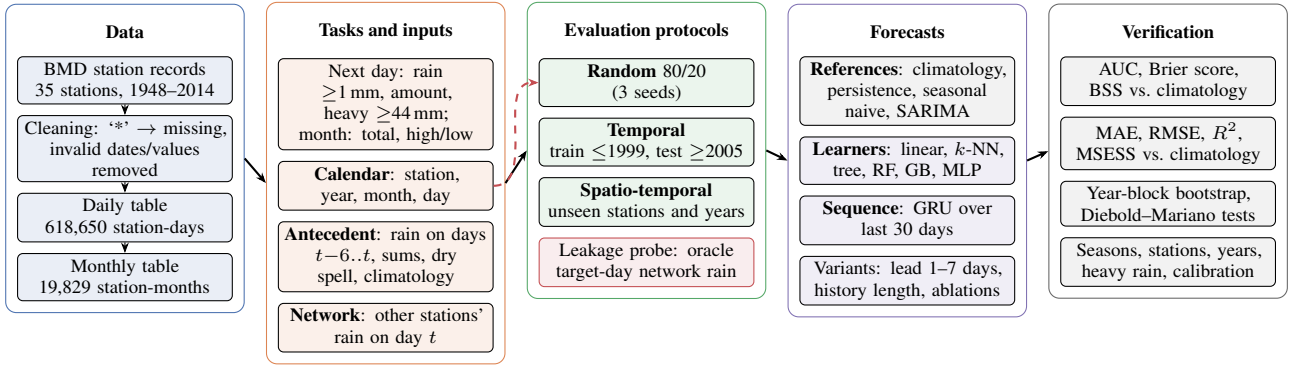
\begin{figure*}[t]
\centering
\definecolor{cdata}{HTML}{4C72B0}\definecolor{cfeat}{HTML}{DD8452}\definecolor{cprot}{HTML}{55A868}\definecolor{cmod}{HTML}{8172B3}\definecolor{cleak}{HTML}{C44E52}
\begin{tikzpicture}[font=\scriptsize, node distance=1.4mm and 4mm,
  box/.style={draw, rounded corners=2pt, align=center, inner sep=2pt, minimum height=5.5mm, text width=#1},
  box/.default=27mm,
  hdr/.style={font=\scriptsize\bfseries, align=center},
  grp/.style={draw, rounded corners=4pt, inner sep=1.5mm},
  arr/.style={-{Stealth[length=1.4mm]}, thin},
  leak/.style={-{Stealth[length=1.4mm]}, thick, dashed, cleak}]
\node[hdr] (h1) at (0mm,0) {Data};
\node[box, fill=cdata!12, below=1.2mm of h1] (raw) {BMD station records\\35 stations, 1948--2014};
\node[box, fill=cdata!12, below=of raw] (clean) {Cleaning: `*' $\rightarrow$ missing,\\invalid dates/values removed};
\node[box, fill=cdata!12, below=of clean] (daily) {Daily table\\618,650 station-days};
\node[box, fill=cdata!12, below=of daily] (month) {Monthly table\\19,829 station-months};
\draw[arr] (raw) -- (clean); \draw[arr] (clean) -- (daily); \draw[arr] (daily) -- (month);
\begin{scope}[on background layer]\node[grp, draw=cdata, fit=(h1)(raw)(month)] (g1) {};\end{scope}
\node[hdr] (h2) at (34.5mm,0) {Tasks and inputs};
\node[box, fill=cfeat!12, below=1.2mm of h2] (tasks) {Next day: rain $\ge$1\,mm, amount,\\heavy $\ge$44\,mm; month: total, high/low};
\node[box, fill=cfeat!12, below=of tasks] (cal) {\textbf{Calendar}: station,\\year, month, day};
\node[box, fill=cfeat!12, below=of cal] (ant) {\textbf{Antecedent}: rain on days\\$t{-}6..t$, sums, dry spell, climatology};
\node[box, fill=cfeat!12, below=of ant] (net) {\textbf{Network}: other stations'\\rain on day $t$};
\begin{scope}[on background layer]\node[grp, draw=cfeat, fit=(h2)(tasks)(net)] (g2) {};\end{scope}
\node[hdr] (h3) at (69mm,0) {Evaluation protocols};
\node[box, fill=cprot!12, below=1.2mm of h3] (rnd) {\textbf{Random} 80/20\\(3 seeds)};
\node[box, fill=cprot!12, below=of rnd] (tmp) {\textbf{Temporal}\\train $\le$1999, test $\ge$2005};
\node[box, fill=cprot!12, below=of tmp] (spt) {\textbf{Spatio-temporal}\\unseen stations and years};
\node[box, fill=cleak!10, draw=cleak, below=of spt] (prb) {Leakage probe: oracle\\target-day network rain};
\begin{scope}[on background layer]\node[grp, draw=cprot, fit=(h3)(rnd)(prb)] (g3) {};\end{scope}
\node[hdr] (h4) at (103.5mm,0) {Forecasts};
\node[box, fill=gray!10, below=1.2mm of h4] (ref) {\textbf{References}: climatology,\\persistence, seasonal naive, SARIMA};
\node[box, fill=cmod!12, below=of ref] (ml) {\textbf{Learners}: linear, $k$-NN,\\tree, RF, GB, MLP};
\node[box, fill=cmod!12, below=of ml] (gru) {\textbf{Sequence}: GRU over\\last 30 days};
\node[box, fill=cmod!12, below=of gru] (vars) {Variants: lead 1--7 days,\\history length, ablations};
\begin{scope}[on background layer]\node[grp, draw=cmod, fit=(h4)(ref)(vars)] (g4) {};\end{scope}
\node[hdr] (h5) at (138mm,0) {Verification};
\node[box, fill=gray!10, below=1.2mm of h5] (prob) {AUC, Brier score,\\BSS vs.\ climatology};
\node[box, fill=gray!10, below=of prob] (amt) {MAE, RMSE, $R^2$,\\MSESS vs.\ climatology};
\node[box, fill=gray!10, below=of amt] (stat) {Year-block bootstrap,\\Diebold--Mariano tests};
\node[box, fill=gray!10, below=of stat] (brk) {Seasons, stations, years,\\heavy rain, calibration};
\begin{scope}[on background layer]\node[grp, draw=black!60, fit=(h5)(prob)(brk)] (g5) {};\end{scope}
\draw[arr, thick] (g1.east) -- (g2.west);
\draw[arr, thick] (g2.east) -- (g3.west);
\draw[arr, thick] (g3.east) -- (g4.west);
\draw[arr, thick] (g4.east) -- (g5.west);
\draw[leak] (cal.east) to[out=0, in=180] (rnd.west);
\end{tikzpicture}
\vspace{-1mm}
\caption{\textbf{Workflow of the study.} Raw BMD records are cleaned so that missing days remain missing, then turned into daily and monthly forecasting tasks. Each model is trained with calendar, antecedent or network inputs and evaluated under three protocols, always against reference forecasts and with skill scores relative to climatology. The dashed red path marks the combination that inflates reported accuracy: calendar inputs with a random split, which lets a model recover the target day's rainfall from neighbouring stations.}
\label{fig:workflow}
\vspace{-2mm}
\end{figure*}

\textbf{Contributions.}
(1) We build a cleaned version of the public BMD station dataset in which missing days are treated as missing rather than as zero rainfall, and define four forecasting tasks: next-day rain occurrence, next-day amount, next-day heavy rain ($\ge$44\,mm), and one-month-ahead totals.
(2) We compare three evaluation protocols (random, temporal, spatio-temporal) and two feature sets (calendar-only, antecedent rainfall) for six learners and four reference forecasts, with repeated seeds, block-bootstrap confidence intervals and Diebold--Mariano tests.
(3) We show that the random split inflates calendar-only models, that climatology explains almost all of the apparent monthly skill, and that lagged rainfall adds real but small next-day skill that varies by season.
(4) In further experiments we show that station and year inputs reduce skill even under a clean split, that skill vanishes beyond two days, that observations from neighbouring stations add skill legitimately, that a recurrent sequence model brings no gain over trees, and that zero-filling missing data creates artificial droughts.
(5) We release code and a checklist for leakage-aware evaluation of station-based rainfall models. \Cref{fig:workflow} summarises the workflow.

\section{Related Work}
\label{sec:related}

\textbf{Station-based machine learning for rainfall.} Many studies treat rainfall forecasting as tabular regression or classification over station records. \citet{cramer2017extensive} compared seven learners for rainfall in weather-derivative pricing and found that the best method depended on the location and that simple models were often competitive. Studies in Malaysia \citep{ridwan2021rainfall}, Vietnam \citep{pham2020daily}, Ethiopia \citep{liyew2021daily}, Pakistan \citep{rahman2022fusion} and the United Kingdom \citep{barrera2022rainfall} compared tree ensembles, support vector machines and neural networks, while recurrent models have been used for monthly rainfall \citep{chhetri2020blstm} and rainfall-runoff modelling \citep{kratzert2018rainfall}. These studies differ in data, target and split, and few report skill against climatology, which makes it difficult to compare them or to know whether a reported accuracy reflects forecasting skill or seasonality.

\textbf{Deep learning for weather and precipitation.} At the other end of the spectrum, deep networks now forecast precipitation from radar and satellite fields. ConvLSTM introduced a spatio-temporal recurrent architecture for nowcasting \citep{shi2015convlstm}, deep generative models produced skilful radar nowcasts that were preferred by forecasters \citep{ravuri2021dgmr}, and MetNet-2 extended neural precipitation forecasts to twelve hours \citep{espeholt2022metnet2}. Global models trained on ERA5 reanalysis \citep{hersbach2020era5} now rival operational numerical weather prediction at medium range \citep{lam2023graphcast,bi2023pangu}, including probabilistic ensembles \citep{price2025gencast}, hybrid physics--ML models \citep{kochkov2024neuralgcm} and foundation models \citep{nguyen2023climax}. Super-resolution and process emulation have also been studied \citep{adewoyin2021trunet,gettelman2021warm}. A recurring lesson from this literature is the importance of standard benchmarks, held-out years and strong reference forecasts \citep{rasp2020weatherbench,schultz2021can,reichstein2019deep}. Station-level studies in data-sparse regions rarely follow the same practice; our work brings it to that setting.

\textbf{Evaluation, leakage and forecast verification.} Leakage occurs when information that would not be available at prediction time enters training or evaluation \citep{kaufman2012leakage}; \citet{kapoor2023leakage} found it in hundreds of ML-based scientific papers, and \citet{lones2024avoid} lists it among the most common pitfalls. For time series, random cross-validation can be optimistic and blocked or forward-chaining schemes are preferred \citep{bergmeir2012use,cerqueira2020evaluating,hyndman2021fpp3}. For spatial data, random splits overstate performance compared with spatial blocking \citep{roberts2017cross,ploton2020spatial,meyer2021aoa}. In meteorology, forecasts are verified with proper scores such as the Brier score \citep{brier1950verification,gneiting2007strictly} and reported as skill against climatology or persistence \citep{murphy1988skill,murphy1993good,wilks2019statistical}. Classical statistical models such as weather generators \citep{richardson1981stochastic}, downscaling regressions \citep{wilby1998statistical} and seasonal ARIMA \citep{box2015time} provide strong baselines, and forecasting competitions show that simple methods are hard to beat \citep{makridakis2020m4}. Paired accuracy differences can be tested with the Diebold--Mariano test and its small-sample correction \citep{diebold1995comparing,harvey1997testing}. We apply these established practices to a public rainfall dataset that has mostly been studied with random splits.

\section{Data and Tasks}
\label{sec:data}

\begin{figure}[t]
  \centering
  \includegraphics[width=\linewidth]{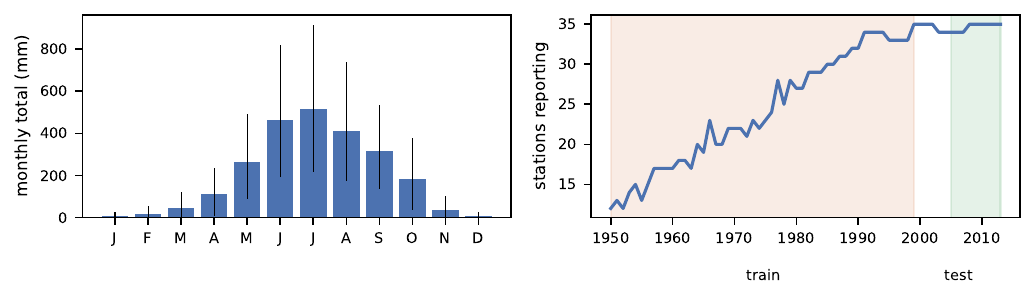}
  \vspace{-6mm}
  \caption{\textbf{Data.} Left: mean monthly total across stations (bars) with 10th--90th percentile range; over 70\% of annual rainfall falls in June--September. Right: number of stations with a valid monthly total per year, with the training (1950--1999) and test (2005--2013) periods shaded.}
  \label{fig:data}
  \vspace{-3mm}
\end{figure}

\textbf{Source.} We use the public ``Historical Rainfall Data in Bangladesh'' collection of BMD station records \citep{redikod_kaggle}. The raw file contains daily rainfall (mm) for 35 stations from January 1948 to June 2014, with one row per station-month and one column per day. Stations enter the network at different times, so the number of reporting stations grows from a handful in the 1950s to 35 by the 1980s (\cref{fig:data}, right).

\textbf{Cleaning.} The raw file marks missing days with asterisks. Derived versions of the file that circulate with the dataset replace these marks with zero, which silently turns missing observations into dry days. We instead treat them as missing (2.17\% of station-days), drop impossible dates (e.g.\ 31 February), and treat negative or implausible values ($>$1000\,mm/day) as missing. The result is 632,377 station-days. Monthly totals are computed only for months with at least 25 observed days and are scaled to the full month length; this gives 19,829 valid station-months (1950--2013).

\textbf{Tasks.} All forecasts are issued at the end of day $t$ (or month $m$) for station $s$ and use only information available at that time.
\begin{itemize}
  \setlength\itemsep{0pt}
  \item \textbf{Next-day occurrence:} is rainfall on day $t{+}1$ at least 1\,mm? (31\% of station-days are wet.)
  \item \textbf{Next-day amount:} rainfall on day $t{+}1$ in mm.
  \item \textbf{Next-day heavy rain:} is rainfall on day $t{+}1$ at least 44\,mm, BMD's threshold for moderately heavy rain? (4.0\% of test days.)
  \item \textbf{Monthly totals:} total rainfall in month $m{+}1$ (regression), and whether it exceeds the 62.5th percentile of training-period totals (203\,mm), so that about three in eight months are labelled high.
\end{itemize}
The daily table has 618,650 station-days with a known next-day value.

\section{Experimental Protocol}
\label{sec:protocol}

\textbf{Evaluation protocols.} We compare three ways to split the data.
\emph{Random}: station-days (or station-months) are assigned to training (80\%) or test (20\%) at random, as is common in prior work; we repeat this with three seeds.
\emph{Temporal}: training uses 1948--1999 and testing uses 2005 to the end of the record (120,172 test station-days; 3,743 test station-months for 2005--2013). The five-year gap limits dependence between the last training days and the first test days.
\emph{Spatio-temporal}: stations are randomly partitioned into five groups; each group is held out in turn, models are trained on the other stations in 1948--1999 and tested on the held-out stations in 2005--2014. This measures how well a model transfers to a location it has never seen, which matters for the many places in Bangladesh without a station.

\textbf{Feature sets.} \emph{Calendar} features (station code, year, month, day) reproduce the common station-and-date setup described above. \emph{Antecedent} features add information that is actually available at forecast time: rainfall on the current and previous six days, 3-, 7- and 30-day totals, the number of wet days in the past week, the length of the current dry spell, a cyclical encoding of day-of-year, and the station's climatological mean and wet-day frequency for that day of year. For monthly forecasts the antecedent features are the totals of the previous 1, 2, 3 and 12 months, a 3-month mean and the station-month climatology. Climatologies are always computed from training rows only; for stations held out in the spatio-temporal protocol we use the network-wide climatology of the training stations.

\textbf{Reference forecasts.} \emph{Climatology} predicts, for each station and day of year, the training-period mean amount and wet-day frequency, smoothed with a $\pm$15-day window (station-month means for monthly tasks). \emph{Persistence} predicts that tomorrow equals today (or next month equals this month). \emph{Seasonal naive} predicts the value from twelve months earlier. \emph{SARIMA}$(1,0,0)(1,1,1)_{12}$ is fitted per station on log-transformed monthly totals for 1950--1999 and used for one-step forecasts with fixed parameters \citep{box2015time,seabold2010statsmodels}; three stations with too short a training record are excluded, so SARIMA is compared with the other methods on the common subset.

\textbf{Learners.} We use logistic or ridge regression, $k$-nearest neighbours ($k{=}50$, fitted on a random 100,000-row subsample for speed), a decision tree (depth 12), a random forest \citep{breiman2001random} (100 trees on daily data, 300 on monthly), histogram gradient boosting \citep{friedman2001greedy,ke2017lightgbm} and a two-layer perceptron (64 units each), all from scikit-learn \citep{pedregosa2011scikit}. In \cref{sec:further} we also train a GRU sequence model (\cref{app:details}). Hyperparameters were fixed in advance at common defaults rather than tuned on the test period. For occurrence, the decision threshold maximises F1 on a 50,000-row sample of the training data.

\textbf{Metrics and tests.} For occurrence we report ROC AUC, the Brier score and the Brier skill score $\mathrm{BSS}=1-\mathrm{BS}/\mathrm{BS}_{\mathrm{clim}}$ \citep{brier1950verification}, plus accuracy, precision, recall and F1 at the chosen threshold. For amounts we report MAE, RMSE, $R^2$ and the MSE skill score $\mathrm{MSESS}=1-\mathrm{MSE}/\mathrm{MSE}_{\mathrm{clim}}$ \citep{murphy1988skill}. A positive skill score means the model beats climatology. Confidence intervals for daily skill scores come from 500 bootstrap resamples of whole test years, which preserves within-year dependence. For monthly totals we apply the Diebold--Mariano test with the Harvey--Leybourne--Newbold correction to the monthly mean absolute-error differential across stations \citep{diebold1995comparing,harvey1997testing}.

\section{Implementation Details}
\label{app:details}

\textbf{Data processing.} Daily values are read from the raw BMD file; entries consisting of asterisks or blanks are treated as missing, impossible dates are removed, and values below 0 or above 1000\,mm are set to missing. For each station the series is re-indexed to a complete daily calendar, so that lagged features always refer to the correct day; a lagged value is missing if the corresponding day is missing. Rolling totals require at least two of three, five of seven and twenty of thirty valid days. The dry-spell length counts consecutive days with less than 1\,mm up to day $t$.

\textbf{Climatology.} For each station and day of year, the mean next-day amount and wet-day frequency are computed from training rows only and smoothed with a 31-day moving window that wraps around the year. For stations absent from the training set (spatio-temporal protocol), the network-wide climatology of the training stations is used. For lead $h$, the climatology is computed for the rainfall $h$ days after each day of year.

\textbf{Learners.} All hyperparameters were fixed before seeing test results (\cref{tab:hyper}). The occurrence threshold used for accuracy, precision, recall and F1 maximises F1 on a 50,000-row sample of the training data over a grid from 0.1 to 0.9. Probabilistic scores (AUC, Brier) do not depend on this threshold.

\begin{table*}[t]
\centering
\caption{Hyperparameters (fixed in advance).}
\label{tab:hyper}
\small
\begin{tabular}{lp{0.72\linewidth}}
\toprule
Model & Setting\\
\midrule
Logistic / ridge regression & median imputation, standardisation; L2 penalty, $C=1$ / $\alpha=1$\\
$k$-nearest neighbours & $k=50$, standardised inputs, fitted on 100,000 random training rows\\
Decision tree & maximum depth 12, at least 20 samples per leaf\\
Random forest & daily: 100 trees, 20\% rows per tree, 50\% features per split; monthly: 300 trees; at least 5 samples per leaf\\
Histogram gradient boosting & 300 iterations, learning rate 0.05; early stopping on 10\% of training rows in the main table, none in \cref{sec:further}\\
MLP & two hidden layers of 64 ReLU units, Adam, early stopping, at most 50 epochs\\
GRU & 30-day input of $\log(1+\text{rain})$ with a missing-day mask; 64 hidden units; 8-dimensional station embedding; seasonal encoding and climatology as static inputs; joint logistic and log-amount heads; Adam, learning rate $10^{-3}$, batch 1024, early stopping on 1995--1999 (patience 3); amounts rescaled by a factor fitted on 1995--1999\\
SARIMA & $(1,0,0)(1,1,1)_{12}$ on $\log(1+\text{total})$, fitted on 1950--1999, parameters fixed for one-step forecasts\\
\bottomrule
\end{tabular}
\end{table*}

\textbf{Statistical tests.} Confidence intervals for daily skill resample the ten test years with replacement (500 resamples) and recompute the score on the pooled resample. The Diebold--Mariano statistic uses the series of monthly mean differences in absolute error across stations, with the Harvey--Leybourne--Newbold small-sample correction and a Student-$t$ reference distribution with $n-1$ degrees of freedom.

\section{Results}
\label{sec:results}

\subsection{Monthly forecasts: how much is seasonality?}
\label{sec:revisit}

\Cref{tab:revisit} reports the classification of high- versus low-rainfall months under random and temporal evaluation.

\begin{table}[t]
\centering
\caption{\textbf{High- vs.\ low-rainfall month classification} (threshold 203\,mm). Random-split values are means over three seeds (s.d.\ $\le$0.8 points). Acc.\ and F1 in \%. Features: Cal = calendar only, Ant = antecedent rainfall.}
\label{tab:revisit}
\small
\setlength{\tabcolsep}{3.5pt}
\begin{tabular}{llcccccc}
\toprule
 & & \multicolumn{3}{c}{Random split} & \multicolumn{3}{c}{Temporal split}\\
\cmidrule(lr){3-5}\cmidrule(lr){6-8}
Model & Feat. & Acc. & AUC & F1 & Acc. & AUC & F1\\
\midrule
Climatology & -- & 87.8 & 0.946 & 84.6 & 88.4 & 0.950 & 84.9\\
Logistic & Cal & 58.9 & 0.609 & 0.0 & 49.1 & 0.607 & 0.1\\
Random forest & Cal & 88.2 & 0.950 & 84.9 & 87.2 & 0.942 & 85.0\\
Grad.\ boosting & Cal & 89.3 & 0.958 & 86.5 & 87.6 & 0.951 & 85.5\\
Logistic & Ant & 88.0 & 0.950 & 84.5 & 88.8 & 0.956 & 85.6\\
Random forest & Ant & 88.6 & 0.956 & 85.5 & 89.2 & 0.956 & 86.5\\
Grad.\ boosting & Ant & 88.7 & 0.956 & 85.6 & 88.4 & 0.953 & 86.0\\
\bottomrule
\end{tabular}
\vspace{-3mm}
\end{table}

Three points stand out. First, the climatology baseline is already very strong: forecasting the training-period frequency of a high month for each station and calendar month gives 88.4\% accuracy and AUC 0.950 on future years. Under temporal evaluation every learner is within 0.8 points of this, and calendar-only models fall below it (87.2--87.6\% accuracy); under the random split the best learner gains 1.5 points. High classification accuracy on this task is therefore explained mostly by the seasonal cycle rather than by the learner. Second, for monthly totals, the calendar-only random forest has a lower MAE than climatology under the random split (77.3 vs.\ 83.9\,mm, mean of three seeds, when both are fitted on the same random training rows), but under temporal evaluation it is worse (87.1 vs.\ 80.3\,mm). The random split rewards the model for memorising which station-months were wet in a given year, information that is not available when forecasting. Third, no regression model beats climatology on monthly totals under temporal evaluation. The zero-filled versions of the file also bias monthly totals and errors downward, so error values computed from them are not comparable with ours.

\begin{table}[t]
\centering
\caption{\textbf{One-month-ahead station totals} (mm). Temporal split: train 1950--1999, test 2005--2013 (3,743 station-months). $^\dagger$SARIMA covers 32 stations (3,419 station-months); on that subset climatology has MAE 79.2 and RMSE 126.7. Random split: mean over three seeds, with climatology refitted on the same random training rows.}
\label{tab:monthly}
\small
\setlength{\tabcolsep}{4pt}
\begin{tabular}{llccc}
\toprule
Model & Feat. & MAE & RMSE & $R^2$\\
\midrule
\multicolumn{5}{l}{\emph{Temporal split}}\\
Climatology & -- & 80.3 & 129.1 & 0.718\\
Persistence & -- & 132.7 & 212.7 & 0.236\\
Seasonal naive & -- & 108.6 & 182.4 & 0.436\\
SARIMA$^\dagger$ & -- & 76.4 & 132.2 & 0.708\\
Ridge & Cal & 186.3 & 241.2 & 0.016\\
Random forest & Cal & 87.1 & 136.9 & 0.683\\
Grad.\ boosting & Cal & 87.3 & 134.6 & 0.694\\
Ridge & Ant & 83.3 & 130.1 & 0.714\\
Random forest & Ant & 85.1 & 133.9 & 0.697\\
Grad.\ boosting & Ant & 89.7 & 138.2 & 0.677\\
\midrule
\multicolumn{5}{l}{\emph{Random split}}\\
Climatology & -- & 83.9 & -- & 0.694\\
Random forest & Cal & 77.3 & -- & 0.718\\
\bottomrule
\end{tabular}
\vspace{-3mm}
\end{table}

\Cref{tab:monthly} shows the full monthly regression comparison. On the 3,419 test station-months where all methods are available, ridge regression and the random forest had significantly larger absolute errors than climatology (Diebold--Mariano $p<10^{-6}$ and $p=0.004$). SARIMA had a slightly lower MAE (76.4 vs.\ 79.2\,mm) but a higher RMSE (132.2 vs.\ 126.7\,mm), and the MAE difference was not significant ($p=0.17$). Persistence and seasonal naive forecasts were far worse than climatology, as expected for a strongly seasonal but noisy variable.

\subsection{Why random splits leak}
\label{sec:leak}

\begin{table}[t]
\centering
\caption{\textbf{Leakage probe} (next day, gradient boosting). ``+ network adds the oracle mean target-day rainfall at all other stations, information a random split exposes implicitly.}
\label{tab:probe}
\small
\begin{tabular}{llccc}
\toprule
Split & Features & AUC & MAE & $R^2$\\
\midrule
Random & Cal & 0.867 & 7.86 & 0.164\\
Random & Cal + network & 0.916 & 6.33 & 0.343\\
Temporal & Cal & 0.849 & 8.34 & 0.081\\
Temporal & Cal + network & 0.921 & 6.13 & 0.366\\
\bottomrule
\end{tabular}
\vspace{-3mm}
\end{table}

The leak is specific to rainfall's spatial structure. On a given day, wet conditions usually extend over many stations (median pairwise correlation 0.31 for amounts and 0.51 for occurrence). Under a random split, about 80\% of the other stations' records for the forecast day are in the training set, so a flexible model with station and date inputs can learn the ``weather of the day''. To make this explicit, \cref{tab:probe} adds the mean target-day rainfall at all other stations as an oracle feature. This feature raises next-day occurrence AUC from 0.849 to 0.921 and amount $R^2$ from 0.08 to 0.37 under temporal evaluation. A random split gives calendar-only models part of this benefit implicitly: their AUC rises from 0.849 (temporal) to 0.867 (random), and $R^2$ doubles from 0.08 to 0.16. Tree models, which can carve out individual dates, benefit most (\cref{sec:daily}).

\begin{figure}[t]
  \centering
  \includegraphics[width=\linewidth]{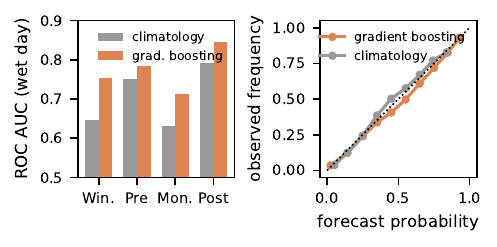}
  \vspace{-6mm}
  \caption{\textbf{Temporal evaluation, gradient boosting with antecedent features.} Left: next-day occurrence AUC by season (winter, pre-monsoon, monsoon, post-monsoon), compared with climatology. Right: reliability diagram; points are 10 probability bins with at least 100 cases.}
  \label{fig:season}
  \vspace{-3mm}
\end{figure}

\subsection{Daily forecasts across protocols}
\label{sec:daily}

\begin{table*}[t]
\centering
\caption{\textbf{Next-day forecasts.} ROC AUC for occurrence ($\ge$1\,mm) and $R^2$ for amount, under random, temporal and spatio-temporal evaluation, with calendar-only (Cal) or antecedent-rainfall (Ant) features. Random split: mean of three seeds (s.d.\ $\le$0.002 AUC). Spatio-temporal: five held-out station groups, weighted by test size. Reference forecasts use no learned features; on held-out stations climatology is the network-wide average. Bold: best per column. Underlined: calendar-only models whose random-split score exceeds the best temporal score with real predictors.}
\label{tab:daily}
\small
\setlength{\tabcolsep}{4pt}
\begin{tabular}{lcccccccccccc}
\toprule
 & \multicolumn{6}{c}{Occurrence: AUC} & \multicolumn{6}{c}{Amount: $R^2$}\\
\cmidrule(lr){2-7}\cmidrule(lr){8-13}
 & \multicolumn{2}{c}{Rand.} & \multicolumn{2}{c}{Temp.} & \multicolumn{2}{c}{Sp.-temp.} & \multicolumn{2}{c}{Rand.} & \multicolumn{2}{c}{Temp.} & \multicolumn{2}{c}{Sp.-temp.}\\
\cmidrule(lr){2-3}\cmidrule(lr){4-5}\cmidrule(lr){6-7}\cmidrule(lr){8-9}\cmidrule(lr){10-11}\cmidrule(lr){12-13}
Model & Cal & Ant & Cal & Ant & Cal & Ant & Cal & Ant & Cal & Ant & Cal & Ant\\
\midrule
Climatology & -- & 0.847 & -- & 0.859 & -- & 0.854 & -- & 0.128 & -- & 0.140 & -- & 0.097\\
Persistence & -- & 0.758 & -- & 0.763 & -- & 0.763 & -- & -0.255 & -- & -0.205 & -- & -0.244\\
\midrule
Logistic / ridge & 0.562 & 0.868 & 0.580 & 0.876 & 0.578 & 0.873 & 0.003 & 0.201 & 0.003 & 0.219 & -0.005 & \textbf{0.184}\\
$k$-NN & 0.843 & 0.865 & 0.843 & 0.870 & \textbf{0.842} & 0.868 & 0.104 & 0.191 & 0.054 & 0.208 & \textbf{0.024} & 0.170\\
Decision tree & 0.870 & 0.872 & 0.817 & 0.862 & 0.812 & 0.852 & 0.152 & 0.177 & 0.003 & 0.157 & -0.111 & 0.108\\
Random forest & \underline{\textbf{0.889}} & 0.883 & 0.835 & 0.879 & 0.832 & 0.877 & \textbf{0.193} & 0.219 & 0.080 & 0.220 & -0.010 & 0.170\\
Grad.\ boosting & 0.866 & \textbf{0.885} & \textbf{0.848} & \textbf{0.879} & 0.839 & 0.877 & 0.165 & \textbf{0.228} & \textbf{0.088} & 0.208 & -0.028 & 0.166\\
MLP & 0.846 & 0.877 & 0.812 & 0.877 & 0.822 & \textbf{0.878} & 0.113 & 0.218 & -0.063 & \textbf{0.235} & 0.001 & 0.183\\
\bottomrule
\end{tabular}
\vspace{-2mm}
\end{table*}

\begin{figure*}[t]
  \centering
  \includegraphics[width=0.92\textwidth]{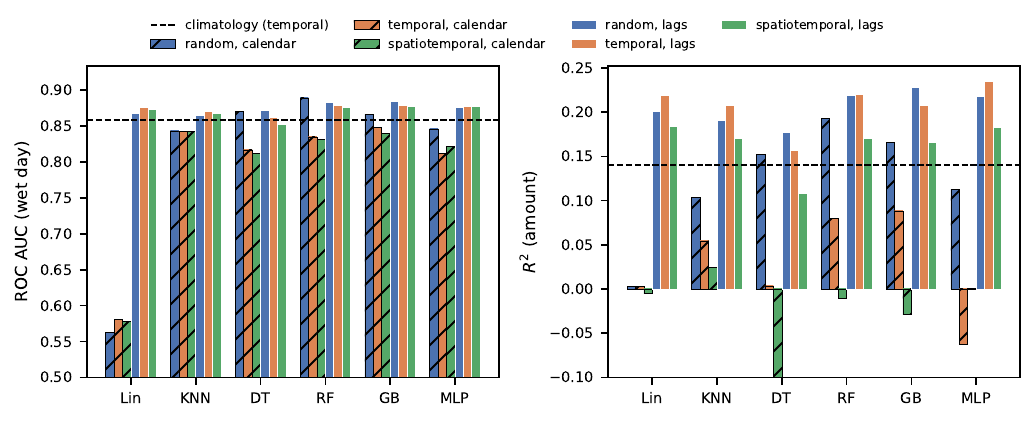}
  \vspace{-3mm}
  \caption{\textbf{Next-day forecasts under three evaluation protocols.} Hatched bars use calendar features only; solid bars add antecedent rainfall. The dashed line is climatology under temporal evaluation. Calendar-only tree models look strong under the random split and fall below climatology on future years and new stations.}
  \label{fig:protocols}
  \vspace{-2mm}
\end{figure*}

\Cref{tab:daily} and \cref{fig:protocols} show next-day results for all learners. Under the random split, the calendar-only random forest reaches an AUC of 0.889, which is higher than every model under temporal evaluation, including those that use the previous week's rainfall (best 0.879). On future years the same model drops to 0.835, below the climatology reference (0.859). The same pattern holds for amounts: the calendar-only random forest has $R^2=0.19$ under the random split but only 0.080 under temporal evaluation and $-0.010$ on unseen stations. A reader who saw only the random-split numbers would conclude that station and date alone predict tomorrow's rain better than recent observations, which is not true.

With antecedent features, the three protocols agree much more closely, because these features do not let the model look up the target day. Under temporal evaluation, the best learners (random forest, gradient boosting, MLP and logistic regression) reach AUC 0.876--0.879 and $R^2$ 0.21--0.24, and differ from each other by less than the gap between them and climatology. Logistic regression is within 0.003 AUC of the best tree ensemble, consistent with findings that the data, not the learner, set the ceiling for tabular problems of this kind \citep{grinsztajn2022tree}. On unseen stations (spatio-temporal protocol), calendar-only models fall to AUC 0.812--0.842, below a network-wide climatology (0.854), because a station code learned on other stations carries no information about a new one. Models with antecedent features lose little: AUC 0.852--0.878 and $R^2$ 0.11--0.18, against 0.854 and 0.097 for climatology. Recent rainfall at the target station is therefore a predictor that transfers across locations, while station and date inputs do not.

\subsection{Skill relative to climatology}
\label{sec:skill}

How much does a model with real predictors add? Under temporal evaluation, gradient boosting with antecedent features has a Brier skill score of 0.074 (95\% CI 0.064--0.088) for occurrence and an MSE skill score of 0.077 (0.064--0.095) for amounts; logistic regression reaches a BSS of 0.063 (0.056--0.069). The AUC gain over climatology is 0.022 (0.018--0.026). These intervals exclude zero, so the skill is real, but it is small: the models remove about 7--8\% of the squared error left by climatology.

Skill varies by season (\cref{fig:season}, left). During the monsoon, climatology alone separates wet and dry days poorly (AUC 0.63) because most days are wet; antecedent rainfall raises AUC to 0.71 but the BSS stays at 0.077. The largest relative gain is in the post-monsoon season (October--November; AUC 0.79 to 0.85, BSS 0.110), when the previous day's rain appears to be especially informative, possibly because post-monsoon rain often comes in multi-day spells. In winter, rain is rare (3.3\% of days) and the BSS is 0.075. Forecast probabilities are well calibrated on future years (\cref{fig:season}, right), which matters more for users than a thresholded accuracy.

\subsection{Heavy rain}
\label{sec:heavy}

Heavy rain ($\ge$44\,mm) occurs on 4.0\% of test days and is the case that matters most for flood warning. A dedicated gradient-boosting classifier reaches AUC 0.866 against 0.825 for a heavy-rain climatology, with a BSS of 0.053 (0.042--0.063). Even with the best possible threshold chosen on the test set, which is optimistic, F1 is only 0.31 (precision 0.27, recall 0.37). Thresholding the regression forecast at 44\,mm detects only 11\% of heavy-rain days, because squared-error regression shrinks toward the mean. Station records alone are therefore not enough for useful heavy-rain warnings at a one-day lead; this is where radar, satellite or numerical weather prediction inputs \citep{ravuri2021dgmr,espeholt2022metnet2,lam2023graphcast} are needed.

\subsection{Further analyses}
\label{sec:further}

\begin{figure*}[t]
  \centering
  \includegraphics[width=\textwidth]{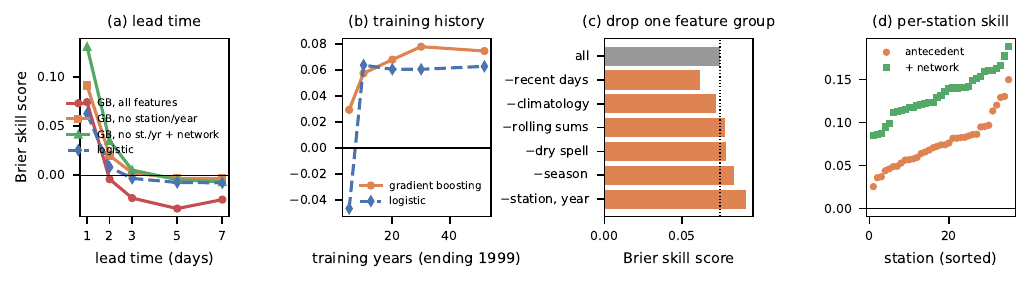}
  \vspace{-6mm}
  \caption{\textbf{Further analyses} (temporal evaluation, next-day occurrence unless stated). (a) Brier skill score against lead time. (b) Skill against the length of the training record, with test years fixed at 2005--2014 and the full-record climatology as reference. (c) Skill after removing one feature group from the gradient-boosting model; the dotted line is the full model. (d) Skill at each of the 35 stations, with and without network observations.}
  \label{fig:more}
  \vspace{-2mm}
\end{figure*}

\begin{table}[t]
\centering
\caption{\textbf{Next-day skill on future years} (temporal split, test 2005--2014). Skill scores are relative to climatology. GB = gradient boosting. ``No st./yr'' drops the station code and year inputs. ``+ network'' adds rainfall observed on day $t$ at the other stations, which is available when the forecast is issued. GRU: mean $\pm$ s.d.\ over three seeds; GB (no st./yr): s.d.\ over five bootstrap refits $\le$0.001.}
\label{tab:skill}
\small
\setlength{\tabcolsep}{3pt}
\begin{tabular}{lccc}
\toprule
Model & AUC & BSS & MSESS\\
\midrule
Climatology & 0.859 & 0 & 0\\
Logistic / ridge (Ant) & 0.876 & 0.063 & 0.092\\
GB (Ant, all inputs) & 0.880 & 0.074 & 0.077\\
GB (Ant, no st./yr) & 0.883 & 0.091 & 0.104\\
GRU (30-day seq.) & 0.884\sd{.000} & 0.092\sd{.001} & 0.095\sd{.003}\\
GB (no st./yr) + network & \textbf{0.896} & \textbf{0.131} & \textbf{0.140}\\
\bottomrule
\end{tabular}
\vspace{-3mm}
\end{table}

\textbf{Station and year inputs hurt.} Removing one feature group at a time (\cref{fig:more}c) shows that the previous seven days of rainfall carry most of the information: dropping them lowers the BSS from 0.074 to 0.061, and dropping the climatology features lowers it to 0.072. Removing the station code and year \emph{raises} the BSS to 0.091 and the MSESS to 0.104 (\cref{tab:skill}); removing the seasonal encoding also helps slightly (0.084). A likely reason is that test years lie outside the training range, so a tree model that splits on year can carry patterns from the last training years into every later year, and a station code lets it fit station-specific noise that the climatology features already summarise. Calendar and identifier inputs, which drive the leakage in \cref{sec:leak}, are therefore harmful even under a clean split.

\textbf{Skill fades after one day.} \Cref{fig:more}a repeats the experiment for forecasts of rain 2, 3, 5 and 7 days ahead, each scored against its own climatology. Skill drops from 0.091 (one day) to 0.020 (two days) and is indistinguishable from zero from three days on. With all inputs, gradient boosting becomes \emph{worse} than climatology at leads of two days or more (BSS between $-0.004$ and $-0.034$), again because of the year and station inputs; without them it is 0.020 at two days and within 0.004 of zero thereafter. Station rainfall carries information about the next day only, which is consistent with the short life of most rain-producing systems; useful forecasts at longer leads require dynamical or large-scale predictors \citep{lam2023graphcast,price2025gencast}.

\textbf{Network observations help legitimately.} Rainfall observed on day $t$ at the other 34 stations is available when the forecast is issued, unlike the target-day values in \cref{tab:probe}. Adding three network summaries (mean rainfall today, mean 3-day total and fraction of wet stations) raises the BSS from 0.091 to 0.131 and the MSESS from 0.104 to 0.140 (\cref{tab:skill}), and, in models that keep all inputs, improves every one of the 35 stations (\cref{fig:more}d; station BSS 0.084--0.188, compared with 0.025--0.150 without network inputs). Skill over climatology is positive at every station and in every test year (BSS 0.054--0.112 per year without network inputs and 0.088--0.157 with them; \cref{tab:year}). The network adds 0.040 BSS at a one-day lead, 0.015 at two days and nothing beyond (\cref{fig:more}a).

\textbf{A sequence model does not change the picture.} A GRU that reads the last 30 days of rainfall with a station embedding and the seasonal context, trained with early stopping on 1995--1999, reaches AUC 0.884, BSS 0.092 and MSESS 0.095, essentially the same as gradient boosting without station and year inputs (\cref{tab:skill}). Ridge regression is also competitive for amounts (MSESS 0.092). The limit is the information in station rainfall, not the model class.

\textbf{How much history is needed?} With the test years fixed, training on only the last 10 years before 2000 already gives a BSS of 0.057 (gradient boosting) and 0.064 (logistic regression), and the gain from 30 or 52 years is small (\cref{fig:more}b). With 5 years, logistic regression is worse than the full-record climatology ($-0.047$). Short records are thus enough for next-day models, but the climatological reference itself benefits from a long record.

\textbf{Missing values are not zeros.} Of the 20,778 station-months in the zero-filled versions of the file, 506 have fewer than 25 observed days and 321 have none. Zero filling turns these months into apparent droughts: their mean total is 28.6\,mm, against a station-month climatology of 184.7\,mm. A further 606 months with a few missing days lose 15.9\,mm on average. Such artificial dry months can dominate error metrics for monthly models and should be removed rather than filled.

\subsection{Additional results}
\label{app:results}

\begin{table*}[t]
\centering
\caption{Next-day occurrence skill by test year (temporal split, gradient boosting with all antecedent inputs, with and without network observations).}
\label{tab:year}
\small
\begin{tabular}{lcccccccccc}
\toprule
Year & 2005 & 2006 & 2007 & 2008 & 2009 & 2010 & 2011 & 2012 & 2013 & 2014\\
\midrule
Wet-day rate & 0.32 & 0.30 & 0.33 & 0.33 & 0.29 & 0.31 & 0.32 & 0.32 & 0.32 & 0.20\\
BSS & 0.112 & 0.054 & 0.081 & 0.056 & 0.054 & 0.060 & 0.082 & 0.068 & 0.091 & 0.092\\
BSS + network & 0.144 & 0.088 & 0.148 & 0.141 & 0.132 & 0.098 & 0.129 & 0.157 & 0.139 & 0.121\\
\bottomrule
\end{tabular}
\end{table*}

\textbf{Per-year skill.} \Cref{tab:year} lists next-day skill for each test year. Skill is positive in every year, ranging from 0.054 to 0.112 with antecedent features and 0.088 to 0.157 with network observations, so the average result is not driven by a single unusual year. The 2014 values cover January--June only.

\textbf{Leakage at the monthly scale.} The monthly random split leaks in the same way as the daily one: other stations' totals for the same month enter the training set. Under the random split, the calendar-only random forest beats a climatology fitted on the same rows by 6.6\,mm MAE on average over three seeds (\cref{tab:monthly}); under the temporal split it is 6.8\,mm worse.

\textbf{Heavy-rain thresholds.} \Cref{sec:heavy} reports F1 at the best threshold chosen on the test set, which is an upper bound. The corresponding probability threshold was 0.18, much lower than 0.5 because heavy rain occurs on only 4\% of days.

\section{Discussion}
\label{sec:discussion}

\textbf{What high accuracy means.} Rainfall categories in Bangladesh can be separated with high accuracy, and with calendar inputs tree ensembles appear far better than linear models. Our analysis shows that this separation comes mostly from seasonality, and that the ranking of learners depends on the split and the features. The near-failure of linear models with calendar inputs (\cref{tab:revisit,tab:daily}) is a property of the feature encoding (month as a number cannot represent a seasonal peak), not evidence that linear models are weak forecasters: with antecedent features, logistic regression is within 0.003 AUC of the best ensemble.

\textbf{A checklist for station-based rainfall studies.} Based on these results, we suggest that studies (i) state the split and use held-out years, with a gap between training and test periods; (ii) use held-out stations when claiming spatial generalisation; (iii) report skill against climatology and persistence, not only accuracy or $R^2$; (iv) use only predictors available at the forecast time and fit climatologies and scalers on training data only; (v) treat missing observations as missing; (vi) report probabilistic scores and calibration for occurrence; (vii) report heavy-rain performance separately; and (viii) avoid station identifiers and year as inputs when forecasting future years. Items (i), (iii), (iv) and (viii) changed our conclusions the most.

\textbf{Limitations.} We use rainfall alone; temperature, humidity, pressure and large-scale indices such as ENSO or reanalysis fields \citep{hersbach2020era5} would likely add skill, especially for monthly forecasts. The station network is sparse and its composition changes over time, so the spatio-temporal protocol tests only transfer between existing stations. Hyperparameters were fixed rather than tuned, which avoids test-set tuning but may leave some performance unused; the narrow spread between very different learners suggests the effect is small. The high/low threshold for monthly classification is a choice; other thresholds would change the absolute accuracies, although the comparison with climatology is the point of interest. Finally, the dataset ends in mid-2014, and more recent BMD data would allow a test on a warmer climate.

\section{Conclusion}
\label{sec:conclusion}

Using six decades of Bangladesh station rainfall, we showed that random train--test splits and calendar-only features make machine-learning rainfall forecasts look much better than they are, and that climatology alone accounts for nearly all of the high accuracy in classifying high- and low-rainfall months. Models that use recent rainfall provide real but modest skill for next-day forecasts, largest after the monsoon and largest overall when same-day observations from the station network are used. Skill disappears beyond two days, is limited for heavy rain, and is not significant for monthly totals. A recurrent sequence model does not change these conclusions. We hope the released protocol helps future studies on data-sparse regions report results that reflect forecasting skill.

\section*{Impact Statement}

This work aims to improve how rainfall forecasting models are evaluated in data-sparse regions. Its main positive impact is to reduce the risk that over-optimistic accuracy figures lead to unwarranted trust in models for flood preparedness or agricultural planning. The models studied here are not operational systems and should not replace official BMD forecasts or warnings. We use only public, non-personal meteorological records.

\section*{Reproducibility}

All code, the cleaned data tables and the raw result files will be released with the paper. Experiments run on a two-core CPU; the complete daily experiment takes about three hours, and the GRU a few minutes per seed. Implementation details are in \cref{app:details}.

\bibliography{refs}
\bibliographystyle{icml2026}

\end{document}